\documentclass[a4paper,fleqn]{cas-sc}

\usepackage{makecell}
\usepackage[authoryear,longnamesfirst]{natbib}
\usepackage{algorithm}
\usepackage{algorithmic}
\def\tsc#1{\csdef{#1}{\textsc{\lowercase{#1}}\xspace}}
\tsc{WGM}
\tsc{QE}

\begin{document}
\let\WriteBookmarks\relax
\def\floatpagepagefraction{1}
\def\textpagefraction{.001}

\shorttitle{SOH-KLSTM Model for Enhanced Lithium-Ion Battery Health Monitoring}    

\shortauthors{}  

\title{SOH-KLSTM: A Hybrid Kolmogorov-Arnold Network and LSTM Model for Enhanced Lithium-Ion Battery Health Monitoring} 

\tnotemark[1] 


\author[first]{Imen Jarraya}
\affiliation[first]{organization={Robotics and Internet of Things Laboratory, College of Computer and Information Sciences},
            city={Riyadh},
            postcode={12435}, 
            country={Saudi Arabia}}

\author[first,second]{Safa Ben Atitallah}
\cormark[1]


 
\author[first]{Fatimah Alahmed}
\author[first]{Mohamed Abdelkader}
\author[first,second]{Maha Driss}
\author[third]{Fatma Abdelhadi}
\affiliation[second]{organization={RIADI Laboratory, National School of Computer Science, University of Manouba}, Department and Organization
          city={Manouba },
          postcode={2010}, 
        country={Tunisia}}
        
\affiliation[third]{organization={College of Electrical and Computer Engineering, King Abdulaziz University},
          city={Jeddah},
          postcode={22254}, 
        country={Saudi Arabia}}
\author[first]{Anis Koubaa}





\begin{abstract}
Accurate and reliable State Of Health (SOH) estimation for Lithium (Li) batteries is critical to ensure the longevity, safety, and optimal performance of applications like electric vehicles, unmanned aerial vehicles, consumer electronics, and renewable energy storage systems. Conventional SOH estimation techniques fail to represent the non-linear and temporal aspects of battery degradation effectively. In this study, we propose a novel SOH prediction framework (SOH-KLSTM) using Kolmogorov-Arnold Network (KAN)-Integrated Candidate Cell State in LSTM for Li batteries Health Monitoring. This hybrid approach combines the ability of LSTM to learn long-term dependencies for accurate time series predictions with KAN's non-linear approximation capabilities to effectively capture complex degradation behaviors in Lithium batteries. KAN addresses LSTM’s limitations in handling non-smooth approximations and memory decay over extended sequences. The combination of LSTM and KAN ensures that the model accurately depicts both the time-dependent changes and the complicated non-linearities of battery degradation. Experimental validation was performed on several subsets from the NASA Prognostics Center of Excellence (PCoE) dataset, which includes Li-ion battery data collected during hundreds of charge-discharge cycles under various operating conditions. The proposed model achieved a Root Mean Square Error (RMSE) of 0.001682 in the NASA B0005 subset, significantly outperforming the LSTM-only model, which achieved an RMSE of 0.058334. This corresponds to a 97.12\% reduction in prediction error, reflecting the superior predictive performance of our proposed model, with an accuracy approximately 35 times greater than that of the LSTM model alone. The results of additional NASA PCoE subsets further highlight the superior performance and computational efficiency of the model, positioning it as a promising solution for real-time battery health monitoring and management systems.
\end{abstract}


\maketitle

\section{Introduction}
\textcolor{black}{Lithium (Li) batteries have emerged as a dominant energy storage solution due to their exceptional energy density, prolonged cycle life, fast charging capability, and adaptability across diverse applications, including electric vehicles, renewable energy systems, and portable electronics \cite{gharehghani2024progress,jarraya2019online,abdelhedi2024optimizing}. However, their performance inevitably degrades with time driven by repeated charge and discharge cycles, temperature fluctuations, and ageing effects \cite{li2024lithium, jarrraya2022comparison}. This degradation not only reduces battery efficiency and reliability but also poses significant safety risks, particularly in high-demand applications where performance consistency is critical \cite{GANDOMAN2019113343}, \cite{TIAN2020120813}.  As a result, accurate estimation of the State of Health (SOH) is essential to ensure the longevity and safe operation of Li batteries.}

\textcolor{black}{SOH is a key indicator of the remaining capacity and functional integrity of a battery relative to its initial state. It encompasses key variables such as voltage, current, temperature, and other factors that influence battery performance. By monitoring these parameters, SOH estimation enables early detection of performance deterioration, allowing proactive maintenance and optimized battery utilization \cite{khaleghi2024towards}. }

Reliable SOH prediction is fundamental to Battery Management Systems (BMS), which enable them to monitor performance, prevent failures, and optimize battery usage. In critical applications such as electric vehicles and large-scale energy storage, inaccurate SOH predictions can cause system malfunctions, unplanned downtime, and safety hazards \cite{seol2023}. \textcolor{black}{Therefore, precise SOH estimation not only improves safety and reliability but also enhances sustainability and cost-effectiveness by extending the lifespan of battery-powered systems \cite{zhang2023battery}.}

Over the years, several approaches for estimating the SOH of Li batteries have been proposed. These approaches can be classified into four main categories: experimental, model-based, data-driven, and fusion methods, as shown in Figure \ref{fig:classification}. Below is a brief overview of these methods.

\begin{itemize}
    \item \textbf{Model-based approaches}, such as electrochemical models and Kalman filters, are more computationally efficient and suitable for real-time use, but are highly dependent on detailed knowledge of the internal states of the battery, which may not always be readily accessible \cite{zheng2023}. Methods based on the Kalman filter have been used, such as the adaptive unscented Kalman filter \cite{LIU2022426}. 
    
    \item \textbf{Experimental techniques}, including capacity measurements and electrochemical impedance spectroscopy, offer high accuracy in assessing battery health, but are often impractical for real-time monitoring due to their invasive nature and dependence on specialized equipment \cite{nuroldayeva2023}. 
    
    \item \textbf{Data-driven methods}, driven by advances in Machine Learning (ML), have demonstrated substantial promise in handling large-scale battery data without the need for in-depth knowledge of internal battery mechanisms. Techniques such as SVR, Random Forests and advanced neural networks such as LSTM and Convolutional Neural Networks (CNNs) have shown remarkable success in identifying complex patterns in battery degradation data \cite{yao2023novel,liang2024,liu2020data}. These methods are particularly effective for the prediction of SOH in real time, where adaptability and predictive accuracy are critical for varying operating conditions.

    \item \textbf{Fusion methods}, which combine two or more methods and can include experimental, model-based, or data-driven approaches, have emerged as a comprehensive solution for SOH estimation by capturing multiple aspects of battery degradation. However, despite their promise, these hybrid models often face challenges due to their computational complexity, which can hinder real-time applications and large-scale deployment \cite{noura2020review}. In this category, a hybrid method based on particle swarm optimization and extreme ML has been proposed for enhanced SOH estimation \cite{CHEN2024100151,chen2024state}.
\end{itemize}
\textbf{\begin{figure*}[h] 
\begin{center} 
\includegraphics[width=13.75cm]{ 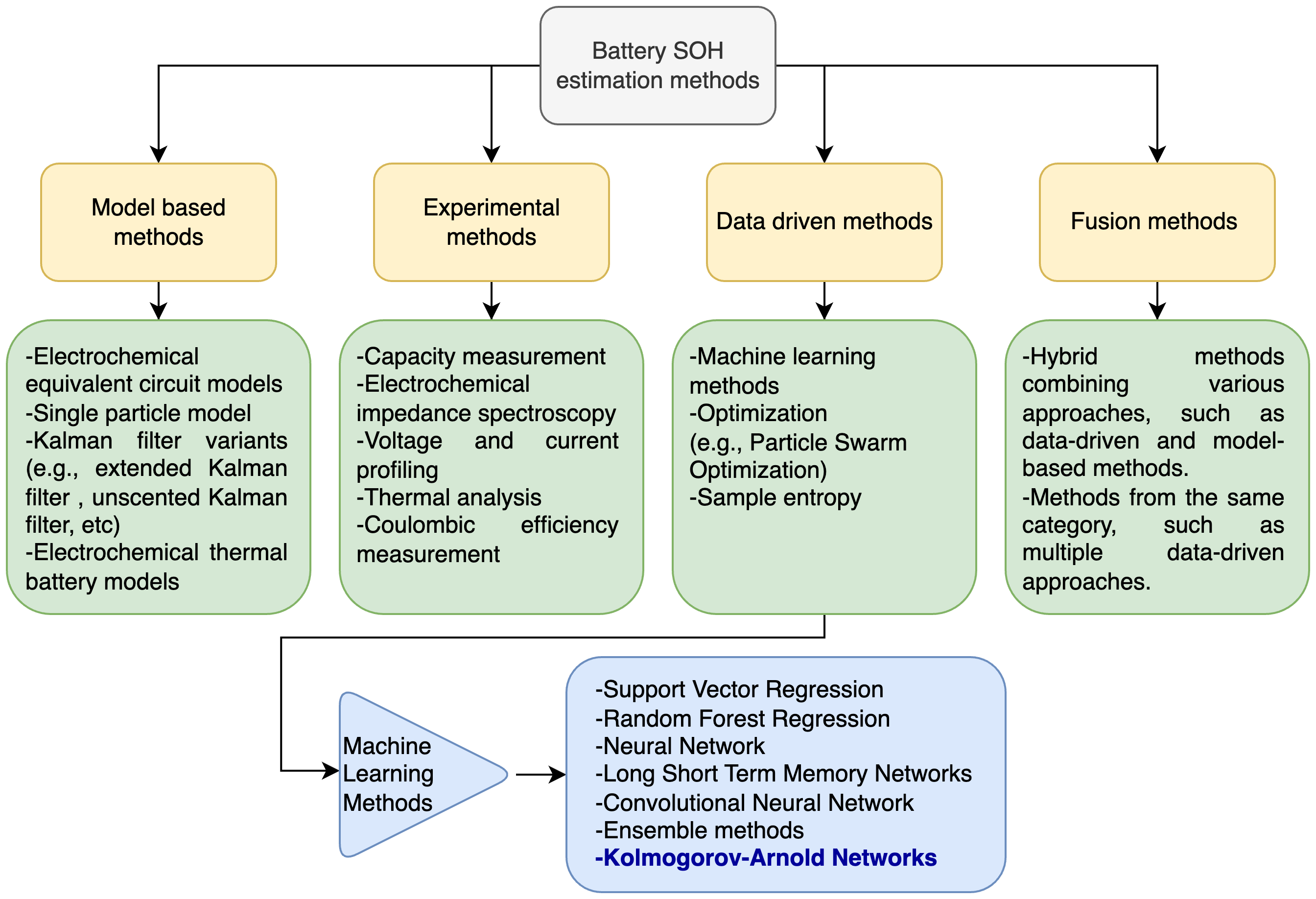} 
\caption{Classification of battery SOH estimation methods.} 
\label{fig:classification} 
\end{center} 
\end{figure*}}

Current SOH estimation models continue to face challenges in accurately representing the complexities associated with battery deterioration. 
\textcolor{black}{These challenges are particularly pronounced in real-world scenarios where batteries operate under diverse environmental conditions, varying load profiles, and across different battery chemistries. The highly complex and nonlinear interactions between internal battery parameters make it difficult for conventional ML models to generalize effectively.}

\textcolor{black}{Recurrent neural networks (RNNs), such as LSTM networks, have shown promise in time-series prediction due to their ability to model sequential dependencies. However, LSTMs still face limitations when handling the non-linear and heterogeneous degradation patterns of lithium-ion batteries. Their reliance on standard activation functions and memory cells restricts their ability to fully capture the intricate relationships that govern battery aging.}

\textcolor{black}{To address these limitations, we introduce a novel hybrid approach, the \textit{SOH-KLSTM} Model, which integrates LSTM networks with the Kolmogorov-Arnold Networks (KAN) to improve the accuracy of SOH prediction. By embedding KAN within the LSTM architecture, we enhance the ability of the model to learn and represent complex, non-linear dependencies in battery degradation data. This integration significantly improves the accuracy of the Li battery health SOH estimation, representing an advance over existing methods that fail to generalize between different battery types and operating conditions. The SOH-KLSTM model is not a simple combination of these two algorithms, but a fundamentally enhanced LSTM design that integrates KAN within the core architecture to improve predictive accuracy. The integration is achieved in an innovative and structurally unique manner:
\begin{itemize}
    \item \textbf{KAN-Enhanced Candidate Cell State}: Conventional LSTM models calculate the potential cell state employing a transformation with fixed weights. Our model replaced this transformation with a KAN-oriented adaptive function that learns non-linear relationships in sequential data dynamically. This enhances the expressiveness of the model, allowing it to capture intricate degradation behaviors more effectively.
    \item \textbf{B-Spline Augmented Feature Space}: Unlike conventional LSTM models that rely just on weight matrices, our approach uses B-spline transformations along with the candidate cell state calculation. This approach allows for the detection of both abrupt and gradual changes in battery degradation trends, thanks to its localized adaptability.
\item \textbf{Self-Learned Activation Functions}: Conventional LSTMs limit adaptability by using predefined activation functions such as sigmoid, tanh, or ReLU. In contrast, our model adapts to the changing dynamics of battery health by dynamically learning activation functions through KAN, which allows us to strengthen the stability of gradient flow.
\end{itemize}
With these changes, the information flow in LSTM networks is much improved, enabling SOH-KLSTM to predict battery SOH while accurately capturing both short- and long-term degradation patterns.
}

The main contributions of this paper are as follows.
\begin{enumerate}
    \item A novel SOH-KLSTM model is introduced that integrates KAN into the LSTM architecture to enhance the prediction of battery SOH by capturing both temporal dependencies and complex non-linear degradation patterns.
    \item The proposed model leverages KAN-enhanced candidate cell state computation, improving the ability of the LSTM to handle intricate degradation behaviors in Li batteries.
    \item The SOH-KLSTM model is implemented and validated using several real-world battery datasets from NASA’s PCoE, demonstrating the superior effectiveness of this hybrid model compared to the standalone LSTM in predicting SOH in diverse operational conditions.
    \item A comparative study highlights the significant improvements of the model over existing SOH estimation methods in terms of predictive accuracy and computational efficiency, demonstrating its superiority for real-time battery management applications.
\end{enumerate}

The structure of this paper is organized to first provide an extensive review of the existing SOH estimation methods in Section \ref{sec:related_work}, covering traditional ML models, hybrid approaches, and data-driven techniques while highlighting the research gaps addressed in this study. The LSTM model used for the estimation of SOH is detailed in Section \ref{sec:battery_SOH}, followed by the introduction of the proposed SOH-KLSTM model in Section \ref{sec:LSTM_KAN}, which describes its architecture and key advantages over conventional approaches. The experimental results, which demonstrate the performance of the model in various datasets, are presented in Section \ref{sec:results}. Finally, the key findings are summarized in Section \ref{sec:conclusion}, along with suggestions for future research to further enhance SOH prediction models.

\section{Related Work}
\label{sec:related_work}

The development of reliable SOH prediction methods for Li batteries \textcolor{black}{progressed significantly}, transitioning from standard ML models to more sophisticated deep learning (DL) frameworks. \textcolor{black}{The primary challenge remains to achieve a balance between prediction accuracy, generalization capability, and computational efficiency, particularly in real-time applications.}

\textcolor{black}{The application of Artificial Intelligence (AI) in SOH prediction has evolved significantly, transitioning from traditional regression-based methods to more advanced deep learning frameworks. 
A notable example is the work of Ma et al. (2022) \cite{ma2022novel}, which introduced an enhanced LSTM-based SOH estimation framework that incorporates the extraction of health indicators, the selection of features, and the optimization of hyperparameters. The study identified 15 health indicators from the charging-discharging process, capturing external battery characteristics such as voltage, current, and temperature to model battery aging and enhancing the accuracy and robustness of the prediction of the model. However, the proposed model relies on predefined health indicators, which limit its adaptability in scenarios where novel degradation patterns emerge.}


\textbf{Hybrid ML} models have emerged as a key advance in the prediction of SOH by combining multiple techniques to take advantage of the strengths of each method \cite{severson2019data,das2024analyzing,amiri2024lithium,fan2020novel,zhang2022prognostics,cai2023early,lu2023deep,bao2023hybrid}.
\textcolor{black}{Bao et al. (2023) \cite{bao2023hybrid} introduced a hybrid deep neural network with dimensional attention (CNN-VLSTM-DA) for SOH estimation, integrating a CNN, a multilayer variant LSTM network and a dimensional attention mechanism.  CNN extracts hierarchical features from battery data, while VLSTM, enhanced with peephole connections, refines the ability to capture long-term dependencies. The dimensional attention mechanism further improves feature selection by assigning different weights to each dimension. The model was validated on NASA, CALCE, and Oxford datasets, demonstrating strong performance under diverse charge / discharge conditions. However, the CNN-VLSTM-DA model comes with a high computational complexity that limits its real-time applicability.}
Zhu et al. (2023) \textcolor{black}{\cite{zhu2023state}} introduced a hybrid framework combining CNNs with LSTM, enabling the model to automatically learn from time series data. CNNs effectively capture spatial features, while LSTMs handle temporal dependencies, making this combination highly effective for SOH prediction. However, despite the improvements, the approach faced challenges such as overfitting, particularly when applied to smaller datasets. 
In addition, Obisakin et al. (2022) \textcolor{black}{\cite{obisakin2022state}} proposed a hybrid model that integrates Support Vector Regression (SVR) with LSTM, achieving an RMSE of 0.005 in the B0005 dataset. Although this hybrid model demonstrated improved accuracy, SVR struggled to capture the long-term dependencies crucial for accurate battery health forecasting. \textcolor{black}{These studies demonstrated that LSTM models, while well-suited for time-series data, could further benefit from domain-specific enhancements to address long-term SOH prediction challenges.}

Recent advances have introduced more specialized methods to improve SOH prediction by integrating diverse techniques. For example, Tao et al. (2024) \cite{TAO2024133541} introduced a multiscale data fusion and anti-noise extended LSTM (MSDF-ANELSTM) model to further enhance SOH prediction. In this approach, the feature extraction process is automated by utilizing the Fast Fourier Transform (FFT) to analyze micro-scale data such as current and voltage in the frequency domain, and then Principal Component Analysis (PCA) is applied to reduce feature redundancy and prevent overfitting. Moreover, the hidden layer structure of the LSTM is improved by separating positive and negative correlation gating weights, reducing the model's complexity and improving generalization. Additionally, a novel combination of Extended Kalman Filter (EKF) and Gradient Descent (GD) for weight updating further enhances noise suppression, addressing a common issue in battery data. As a result, this method significantly outperformed traditional LSTM models, achieving a 66.66\% improvement in accuracy, 83.84\% better stability, and 72.54\% improved generalization.

\textcolor{black}{In addition, recent developments in SOH estimation have explored \textbf{model-data fusion} approaches, combining physics-based modeling and DL to improve predictive performance. Chen et al. (2024) \cite{chen2024state} introduced a hybrid SOH estimation framework that integrates a fractional-order RC Equivalent Circuit Model (ECM) with a DL network. Their approach begins with correlation analysis to extract health features from raw battery data, followed by fractional particle swarm optimization. These optimized parameters, which capture the internal dynamics of the battery, are further analyzed to select the most relevant SOH indicators. An improved Vision Transformer (VIT) network is then trained using the selected health features. The experimental results demonstrate that their method achieves higher predictive accuracy than conventional data-driven models. However, this approach introduces additional computational complexity due to ECM parameter identification and feature correlation analysis. } 
Moreover, Wang et al. (2024) \cite{wang2024physics} introduced \textbf{Physics-Informed Neural Networks} (PINN), which combine empirical degradation models and state-space equations for the estimation of SOH. By embedding physics-based principles into neural networks, their model achieved a MAPE of 0.87\% on a dataset of 387 batteries, improving interpretability and model robustness. However, PINNs, despite their precision, can be computationally intensive, limiting their scalability in large-scale real-time applications.

Further innovation came with the introduction of\textbf{ Self-Supervised Learning (SSL) methods }. Che et al. (2023) \cite{che2023boosting} proposed an SSL framework to address the issue of limited labeled data for the prediction of SOH. By combining auto-encoder-decoder architectures with random forest regression, the model effectively learned hidden aging characteristics from unlabeled data, significantly reducing reliance on large labeled datasets. The model achieved robust performance, with an error distribution below 4\% and overall errors less than 1.14\%. However, the ensemble-based approach introduced computational complexity, limiting its suitability for real-time SOH estimation in large-scale systems where efficiency is critical.

Furthermore, \textbf{Graph-based models} provide an innovative approach by capturing the underlying relationships between battery parameters. Yao et al. (2023) \cite{yao2023novel} introduced a novel graph-based framework, CL-GraphSAGE, for the prediction of SOH, which captures both temporal and spatial dependencies in battery health indicators (HIs). The model utilizes Pearson's correlation coefficients to identify HIs highly correlated with SOH, forming a graph structure to enhance prediction accuracy. Temporal features are captured using CNNs and LSTMs, while spatial relationships are modeled through the GraphSAGE architecture, which propagates information through the graph. The results showed that CL-GraphSAGE outperformed traditional methods such as CNN, LSTM, and GCN, achieving an RMSE as low as 0.2\% on datasets from MIT, NASA, and experimental sources. This validation in diverse data sets confirmed the robustness of the model. However, while CL-GraphSAGE improved accuracy by leveraging spatial and temporal data, it struggled to capture sequential dependencies, which limited its effectiveness for long-term SOH prediction.

Similarly, Wei et al. (2024) \cite{wei2024state} proposed the Conditional Graph Convolutional Network (CGCN), designed to enhance SOH and Remaining Useful Life (RUL) predictions by capturing both feature-to-feature and feature-to-SOH correlations. The CGCN framework utilizes two types of undirected graphs: one to model relationships between battery features and another to model the correlations between those features and SOH or RUL. To further refine temporal predictions, the model incorporates dilated convolutional operations that expand the receptive field and improve the capture of long-term dependencies in time-series data without significantly increasing computational complexity. The results demonstrated a notable improvement in predictive accuracy.  CGCN achieved RMSE values of 0.73\% for SOH and 0.92\% for RUL on NASA and Oxford datasets, outperforming traditional GCN and other ML models in these tests. This improvement was attributed to the model's ability to transmit information more effectively through the graph, capturing both temporal and spatial dependencies. However, despite these gains in accuracy, the CGCN faced challenges in generalizing across highly variable operational conditions, a key challenge for scalable SOH and RUL prediction. The performance of the model was particularly dependent on the quality and consistency of the data, which could limit its effectiveness in environments with significant operational variability.

Recent works have also explored \textbf{attention mechanisms} to improve the accuracy of the prediction of SOH \cite{wang2023bioinspired,zhao2023state}. Zhao et al. (2023) \cite{zhao2023state} introduced a Multi-Head Attention-Time Convolution Network (MHAT-TCN), which integrates multi-head attention learning with gray relational analysis (GRA) to identify key health indicators (HIs) correlated with battery capacity. This approach improves the accuracy of SOH prediction by focusing on relevant HIs during the training process. The MHAT-TCN was validated using leave-one-out cross-validation (LOOCV) across datasets from similar battery models. The model demonstrated significant improvements in the prediction of SOH, with RMSE values of 0.0262 and MAPE of 3.6990, outperforming conventional models such as TCN and LSTM. This method improves predictive performance by capturing local regeneration phenomena, a key factor in battery degradation analysis. Although the model showed superior accuracy across various datasets, its increased computational complexity remains a limitation, particularly when applied in real-time applications where faster predictions are necessary.
 
\textcolor{black}{In conclusion, while various SOH prediction methods offer unique strengths, they continue to present trade-offs between accuracy, interpretability, and scalability \cite{khaleghi2024towards,su2024state,hindawi2024overview,dini2024review,xia2024technologies}. Hybrid models, which combine multiple techniques, and graph-based approaches show significant promise due to their ability to capture complex relationships in the data. However, the increasing demand for large-scale real-time applications, particularly in practical battery management systems (BMS), necessitates further advancements to enhance efficiency, reliability, and adaptability. 
A major challenge remains the computational complexity associated with advanced SOH prediction models. Although deep learning, hybrid methods, and attention-based models improve predictive performance, their feasibility for real-time deployment is challenged by high computational demands. Reducing complexity while maintaining predictive accuracy is critical for enabling widespread adoption in industrial and automotive applications. Furthermore, improving generalization capabilities across diverse operational conditions remains a key objective to ensure robust and adaptable performance in real world scenarios.}

One of the main contributions in this paper is the fusion of LSTM networks with KANs \cite{hou2024comprehensive}. In 2024, KANs recently gained significant traction as a promising approach, offering advantages over traditional techniques. 
\textcolor{black}{While some studies have explored their application in SOH estimation \cite{zhang2024advanced,peng2024state,chen2025parallel,zhou2024lithium}, their full potential remains underutilized, particularly in hybrid architectures that integrate sequential learning and nonlinear function approximation.}
To address existing limitations, we propose the SOH-KLSTM model, which integrates the strengths of the LSTM and KAN networks. In fact, KAN excels in capturing the non-linear degradation behaviors of Li batteries, while LSTM is proficient in modeling temporal sequences of battery usage. By combining these two approaches, the SOH-KLSTM model provides a comprehensive solution that improves prediction accuracy and generalization under different battery conditions. This hybrid architecture provides a scalable and adaptable approach for the prediction of SOH with high precision, generalization, and computational efficiency.
\textcolor{black}{ Unlike previous hybrid approaches, such as KAN-LSTM \cite{zhang2024advanced} and CNN-KAN \cite{peng2024state}, where KAN is applied before or after the used model, our approach directly integrates KAN into the computation of the candidate cell state of the LSTM. This integration employs B-spline transformations and SiLU activation functions. In contrast, existing studies have applied KAN differently. Zhang et al. \cite{zhang2024advanced} use KAN for feature compression before passing data to an LSTM. Peng et al. \cite{peng2024state} apply KAN after LSTM to refine the extracted temporal features.
Chen et al. \cite{chen2025parallel} incorporate KAN within a Transformer-based fusion framework, using B-spline interpolation for high-dimensional feature transformations.
Zhang et al. \cite{zhou2024lithium} integrate KAN with CNN, where it converts high-level CNN-extracted features into refined SOH predictions.}

\textcolor{black}{In our approach, by embedding KAN within the LSTM structure at an earlier stage, the SOH-KLSTM model offers; improved feature representation, more fine-grained control over hidden state updates, and better generalization across diverse battery types and operational conditions.}

\section{Battery State of Health Estimation: Methodology}
\label{sec:battery_SOH}

\begin{figure*}[!h]
\begin{center}
\includegraphics[width=1\textwidth]{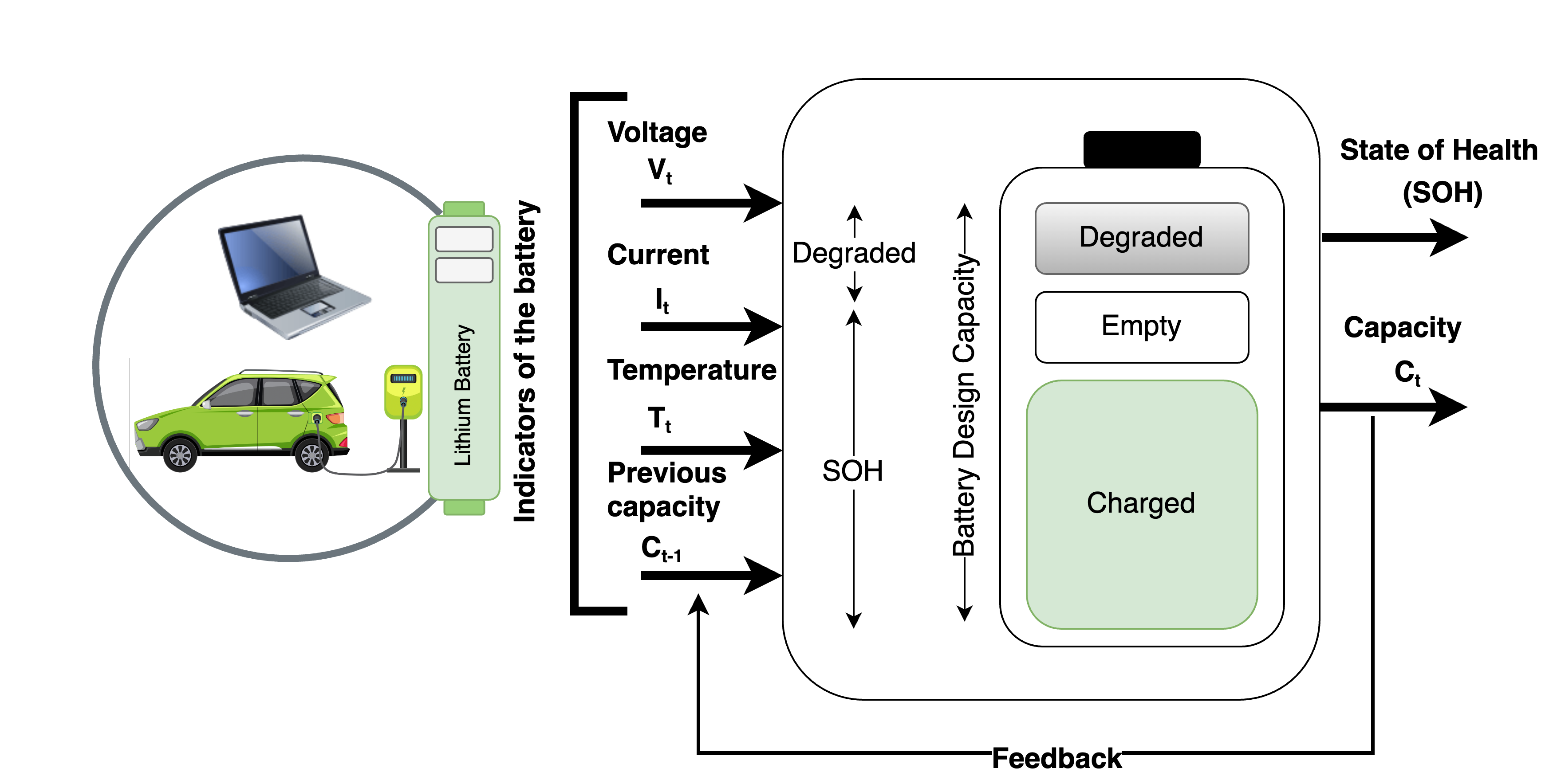}
\caption{
\textcolor{black}{SOH estimation process for a lithium-ion battery. This diagram illustrates the key indicators used in SOH estimation, including voltage $(V_t)$, current $(I_t)$, temperature $(T_t)$, and previous capacity $(C_{t+1})$. These indicators influence battery degradation and overall capacity $(C_t)$, which is monitored to assess SOH.}}
\label{SOH}
\end{center}
\end{figure*}

\textcolor{black}{The health of Li batteries is an important factor in optimizing energy storage investments, reducing maintenance costs, and ensuring reliable operation. Accurate SOH estimation is essential to effectively manage battery performance.} However, this process is complex and influenced by several interrelated factors, such as charge-discharge cycles, environmental conditions, and internal resistance.
These factors significantly affect the degradation rate of the battery, which in turn affects its safety, energy efficiency, and overall useful life. Specifically, key parameters such as temperature variations, current loads, and voltage changes further contribute to battery deterioration, underscoring the need for real-time monitoring to ensure accurate predictions, as shown in Figure~\ref{SOH}.

\subsection{Problem Statement}
The SOH is a critical metric for evaluating a battery's current performance and estimating its remaining lifespan. It represents the residual capacity of the battery relative to its nominal (initial) capacity when new. Typically, SOH is calculated using Equation \ref{eq:1} \cite{liu2023online}:

\begin{equation}
    SOH = \frac{C_t}{C_{nominal}} \times 100\%
    \label{eq:1}
\end{equation}

where \( C_t \) represents the current battery capacity at time \( t \), and \( C_{nominal} \) is the nominal capacity specified by the manufacturer when the battery is new.

In the context of battery health monitoring, particularly for assessing energy efficiency, an alternative expression of SOH is based on the ratio of charge throughput. Equation \ref{eq:2} defines this approach, which computes the ratio of the battery's starting capacity to its total charge provided \cite{liu2023online}:

\begin{equation}
    SOH = \frac{Q_{out}}{Q_{in}} \times 100\%
    \label{eq:2}
\end{equation}

where \( Q_{out} \) is the total discharged energy, and \( Q_{in} \) is the total energy charged into the battery. This formulation highlights the battery's ability to retain energy during operation, which is essential for applications like electric vehicles and energy storage systems.

As the battery undergoes continuous charge and discharge cycles, its capacity \( C_t \) gradually decreases with time, directly influencing the SOH. Modeling this degradation requires capturing intricate temporal dependencies and non-linear relationships among various operational parameters. The proposed SOH-KLSTM model employs a set of time series input features to estimate both the SOH and the remaining battery capacity. These features are integrated into the input matrix \( X_t \), representing the operational state of the system at time \( t \). The model predicts SOH and capacity using a differential equation-based approach as defined in Equation \ref{eq:33}:

\begin{equation}
    \frac{d \hat{y}_{\text{SOH}}}{dt}, \frac{d \hat{y}_{\text{cap}}}{dt} = f\left( X_t, \frac{d h_{t-1}}{dt}, \frac{d C_{t-1}}{dt}; \theta \right)
    \label{eq:33}
\end{equation}

where \( \hat{y}_{\text{SOH}} \) is the predicted SOH, \( \hat{y}_{\text{cap}} \) is the predicted battery capacity, \( X_t \) is the vector of input characteristics at time \( t \), \( h_{t-1} \) and \( C_{t-1} \) denote the hidden and cell states from the previous time step, and \( \theta \) represents the trainable model parameters, such as weights and biases. 
The input matrix \( X_t \) plays an essential role in accurately predicting both SOH and capacity by capturing the necessary operational parameters, including current, voltage, temperature, and capacity of the previous cycle. These features are expressed in Equation \ref{eq:44}:

\begin{equation}
    X_t = 
    \begin{bmatrix}
        \text{C}_{t-1}, V_t, I_t, T_t
    \end{bmatrix}
    \label{eq:44}
\end{equation}

where \( \text{C}_{t-1} \) is the capacity of the previous cycle, \( V_t \) is the terminal voltage at time \( t \), \( I_t \) is the terminal current at time \( t \), \( T_t \) is the battery temperature at time \( t \).  

Each of these parameters plays a distinct role in battery performance.  The capacity \( C_t \) reflects the capacity of the battery to store energy, which decreases with age. Voltage \( V_t \) drops as the internal resistance increases, signaling degradation. Current \( I_t \) influences heat generation and efficiency, while temperature \( T_t \) impacts chemical reactions, accelerating or slowing aging. These operational features enable the SOH-KLSTM model to capture non-linear degradation patterns and predict SOH and capacity with high accuracy over time.
\subsection{LSTM for SOH Estimation}
Recent advancements in ML, particularly long-short-term memory networks \cite{zhang2024prediction, chen2024novel, sun2024novel, xu2023improved, chen2024data, rauf2022machine}, have demonstrated their effectiveness in the prediction of SOH. LSTM models excel at capturing temporal dependencies within sequential data, making them highly suitable for real-time monitoring and identifying nonlinear degradation patterns over time. These characteristics make LSTM networks ideal for battery management systems, as they enable accurate diagnostics, proactive decision-making, and failure prevention.

\subsubsection{LSTM Gate Mechanisms}
LSTM networks process sequential battery data and capture long-term dependencies within time series features \cite{dini2024review, ren2023review}. Their architecture incorporates specialized gating mechanisms, including input, forget, and output gates, to manage the flow of information across time steps. This structure keeps relevant information while discarding irrelevant data, improving the accuracy of the SOH estimation. Managing memory over time is essential for neural networks, and LSTMs achieve this with the following gate mechanisms:
\begin{enumerate}
    \item \textbf{Input Gate (\( i_t \))}: controls the extent to which new information is passed into the cell state at time \( t \):
\begin{equation}
    i_t = \sigma(W_i \cdot [h_{t-1}, X_t] + b_i)
\end{equation}
where \( \sigma \) is the sigmoid activation function, \( W_i \) is the weight matrix for the input gate, \( h_{t-1} \) is the previous hidden state, \( X_t \) is the current input, and \( b_i \) is the bias term for the input gate.

    \item \textbf{Forget Gate (\( f_t \))}: determines how much of the previous cell state should be retained:
\begin{equation}
    f_t = \sigma(W_f \cdot [h_{t-1}, X_t] + b_f)
\end{equation}
where \( W_f \) is the weight matrix for the forget gate, and \( b_f \) is its bias term. 

    \item \textbf{Output Gate (\( o_t \))}: regulates how much of the updated cell state is passed to the hidden state:
\begin{equation}
    o_t = \sigma(W_o \cdot [h_{t-1}, X_t] + b_o)
\end{equation}
where \( W_o \) is the weight matrix for the output gate, and \( b_o \) is the corresponding bias term. 
    
\end{enumerate}

These gate mechanisms manage the network's memory and ensure that the LSTM effectively captures both short-term and long-term dependencies within the data.

\subsubsection{Cell State and Hidden State Updates}
The LSTM model maintains two core components, the cell state and the hidden state, to update and retain information over time. These components ensure that the network dynamically adjusts its memory, retaining relevant information while eliminating unnecessary details.

\begin{enumerate} 
    \item \textbf{Cell State Update}: The new cell state \(C_t\) is calculated by combining the previous cell state \(C_{t-1}\), modulated by the forget gate, with new candidate information regulated by the input gate:
    \begin{equation}
        C_t = f_t \cdot C_{t-1} + i_t \cdot \tilde{C}_t
    \end{equation}
    This mechanism ensures the selective retention of past memory while incorporating new information as needed.

    \item \textbf{Hidden State Update}: The hidden state \(h_t\) is updated based on the new cell state, filtered through the output gate:
    \begin{equation}
        h_t = o_t \cdot \tanh(C_t)
        \label{eq:8}
    \end{equation}
    This update ensures that relevant aspects of the cell state are passed to the hidden state for the next time step, maintaining continuity across the sequence.
\end{enumerate}

These updates allow the LSTM to dynamically adapt its internal memory, retaining essential information over long periods and discarding irrelevant data. This capability makes LSTM networks highly effective in managing time-dependent relationships within battery data, leading to improved battery management performance.
\subsubsection{SOH and Capacity Estimation}
Once the LSTM processes the input sequence, the final hidden state \(h_t\), which captures the temporal dependencies and relevant information from the input features, is passed through a fully connected layer. This layer transforms the hidden state into the final predictions for both the SOH and the battery capacity. The prediction is expressed in Equation \ref{eq:9}
\begin{equation}
    \hat{y}_{\text{SOH}}, \hat{y}_{\text{cap}} = W_{\text{out}} \cdot h_t + b_{\text{out}}
    \label{eq:9}
\end{equation}

where \(W_{\text{out}}\) represents the weight matrix of the fully connected layer, and \(b_{\text{out}}\) is the bias vector that adjusts the predictions. The outputs \(\hat{y}_{\text{SOH}}\) and \(\hat{y}_{\text{cap}}\) correspond to the predicted SOH and battery capacity, respectively.

The overall process of SOH and capacity estimation using the LSTM network is summarized in Algorithm~\ref{alg:lstm_soh_capacity}. The algorithm details how the LSTM processes sequential battery data, updates its internal states, and generates predictions for SOH and battery capacity.
\begin{algorithm}
\caption{LSTM for SOH and Capacity Estimation}
\label{alg:lstm_soh_capacity}
\begin{algorithmic}
\STATE \textbf{Input}: Input sequence \( X_t \), hidden state \( h_{t-1} \), cell state \( C_{t-1} \)
\STATE \textbf{Output}: Predicted SOH \( \hat{y}_{\text{SOH}} \) and capacity \( \hat{y}_{\text{cap}} \)

\STATE \textbf{Initialization}: 
\STATE Initialize weights \( W_i, W_f, W_o, W_C \) and biases \( b_i, b_f, b_o, b_C \)
\STATE Initialize hidden state \( h_0 \) and cell state \( C_0 \)

\FOR{each time step \( t \)}
    \STATE Compute pre-activation: \( z_t = W \cdot X_t + U \cdot h_{t-1} + b \)
    \STATE Update gates:
    \STATE \( i_t = \sigma(z_{t,0}) \) \hfill \textit{(Input gate)}
    \STATE \( f_t = \sigma(z_{t,1}) \) \hfill \textit{(Forget gate)}
    \STATE \( o_t = \sigma(z_{t,3}) \) \hfill \textit{(Output gate)}
    \STATE Compute candidate cell state: \( \tilde{C}_t = \tanh(z_{t,2}) \)
    \STATE Update cell state: \( C_t = f_t \cdot C_{t-1} + i_t \cdot \tilde{C}_t \)
    \STATE Update hidden state: \( h_t = o_t \cdot \tanh(C_t) \)
\ENDFOR

\STATE \textbf{Final SOH and Capacity Estimation}: 
\STATE \( \hat{y}_{\text{SOH}}, \hat{y}_{\text{cap}} = W_{\text{out}} \cdot h_t + b_{\text{out}} \)
\end{algorithmic}
\end{algorithm}

\subsection{Kolmogorov-Arnold Networks}
\begin{figure}[ht]
    \centering
    \includegraphics[width=\linewidth]{ 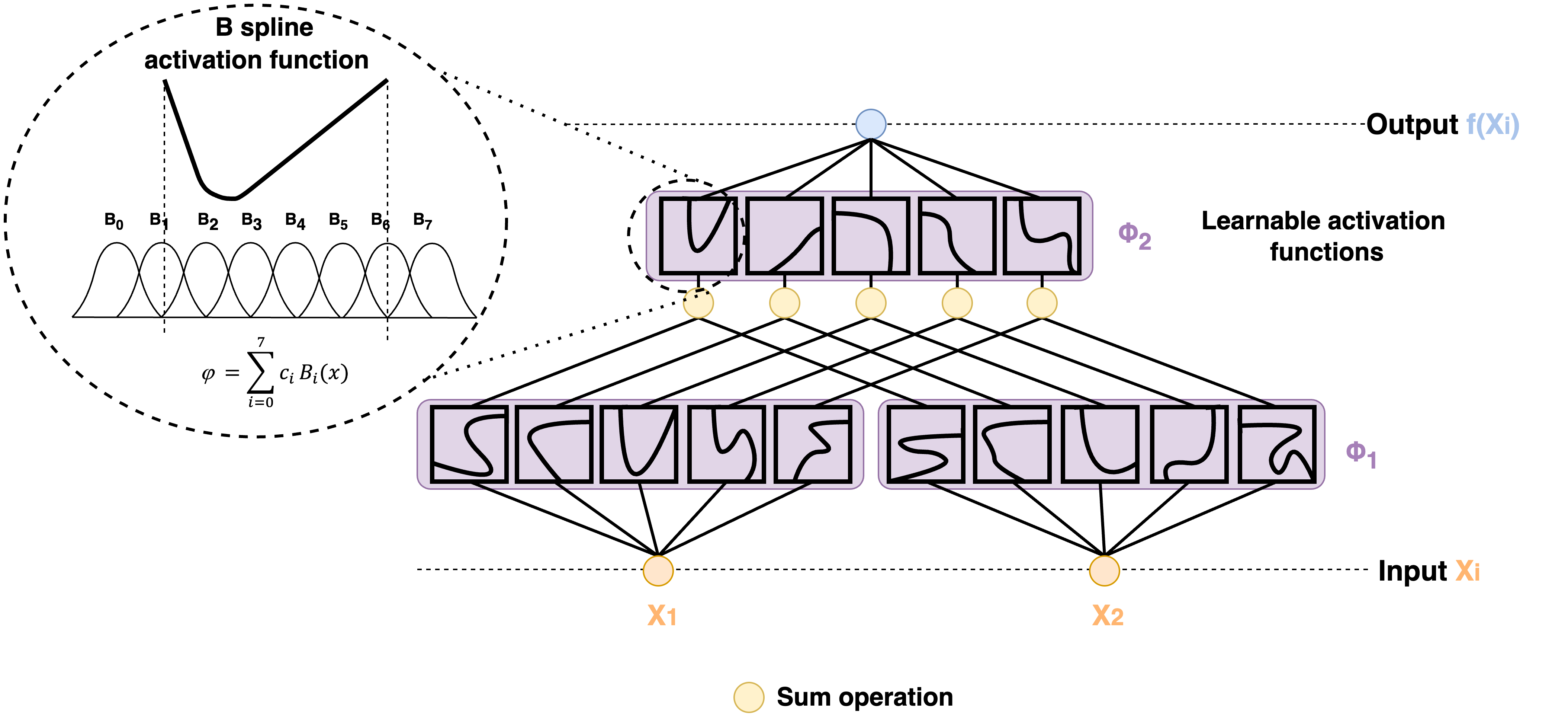}
    \caption{\textcolor{black}{The hierarchical structure of the two-layer KAN approach, where input features $(X_1, X_2)$ are transformed using B-spline functions $(\varphi)$, summed, and processed through learnable activation functions $(\Phi_1, \Phi_2)$. The B-spline activation function inset illustrates the basis expansion process, demonstrating how localized feature representations are combined to enhance predictive performance.} }
    \label{fig:KAN_approach}
\end{figure}

\textcolor{black}{Kolmogorov-Arnold Networks is a universal function approximator that learns an adaptive, non-linear transformation of input features without relying on predefined activation functions.} KAN addresses several limitations inherent in traditional deep learning models, such as Multi-Layer Perceptrons (MLPs) \cite{liu2024kan}. Although MLPs are effective at modeling complex patterns, they often struggle with issues of interpretability and accuracy due to their reliance on fixed activation functions. In contrast, KANs overcome these challenges by employing learnable activation functions along the edges, enabling a more nuanced capture of non-linear dependencies. The theoretical foundation of KANs is the Kolmogorov-Arnold theorem, which guarantees that any continuous multivariate function \( f(x_1, x_2, \dots, x_n) \) can be decomposed into a finite sum of continuous univariate functions. In particular, the theorem asserts that there exist continuous functions \(\Phi_q\) and \(\varphi_{q,p}\) such that:
\begin{equation}
    f(x) = \sum_{q=1}^{2n+1} \Phi_q \left( \sum_{p=1}^{n} \varphi_{q,p}(x_p) \right)
    \label{eq:kan_general}
\end{equation}

where \( \Phi_q \) is responsible for aggregating the transformed inputs, while each \( \varphi_{q,p}(x_p) \) individually transforms the \(p\)-th input feature. The indices \( p \) and \( q \) denote the transformation stages and \( n \) is the total number of input characteristics. This decomposition provides a flexible framework for representing complex functions in high-dimensional spaces, as shown in Figure~\ref{fig:KAN_approach}.

\textcolor{black}{In the KAN framework, each activation function is designed to be learnable and is modeled as a combination of a basis function and a B-spline transformation:}
\begin{equation}
  \textcolor{black}{ \varphi(x) = w_b\, b(x) + w_s\, \text{spline}(x)}
   \label{eq:kan_activation}
\end{equation}
\textcolor{black}{ \( w_b \) and \( w_s \) are trainable scalar weights that control the contributions of the basis function \( b(x) \) and the B-spline transformation \(\text{spline}(x)\), respectively. Although these weights could be merged into the functions \( b(x) \) and \(\text{spline}(x)\), keeping them separate offers finer control over the magnitude of activation and improves training stability. Typically, the basis function \( b(x) \) is implemented using the SiLU activation function, which provides smooth, nonmonotonic transformations and promotes gradient stability.}
\textcolor{black}{The \text{spline}(x)\ transformation is expressed as a linear combination of B-spline basis functions:}
\begin{equation}
  \textcolor{black}{  \text{spline}(x) = \sum_{i} c_i B_i(x)}
    \label{eq:kan_spline}
\end{equation}
\textcolor{black}{where \( B_i(x) \) are the B-spline basis functions and \( c_i \) are trainable coefficients. This representation allows the model to learn localized transformations, which are particularly effective in capturing fine-grained patterns in the data.}

\section{SOH-KLSTM: Proposed KAN-Integrated Candidate Cell State in LSTM Model}
\label{sec:LSTM_KAN}
\begin{figure*}[ht]
    \centering
    \includegraphics[width=0.74\linewidth]{ 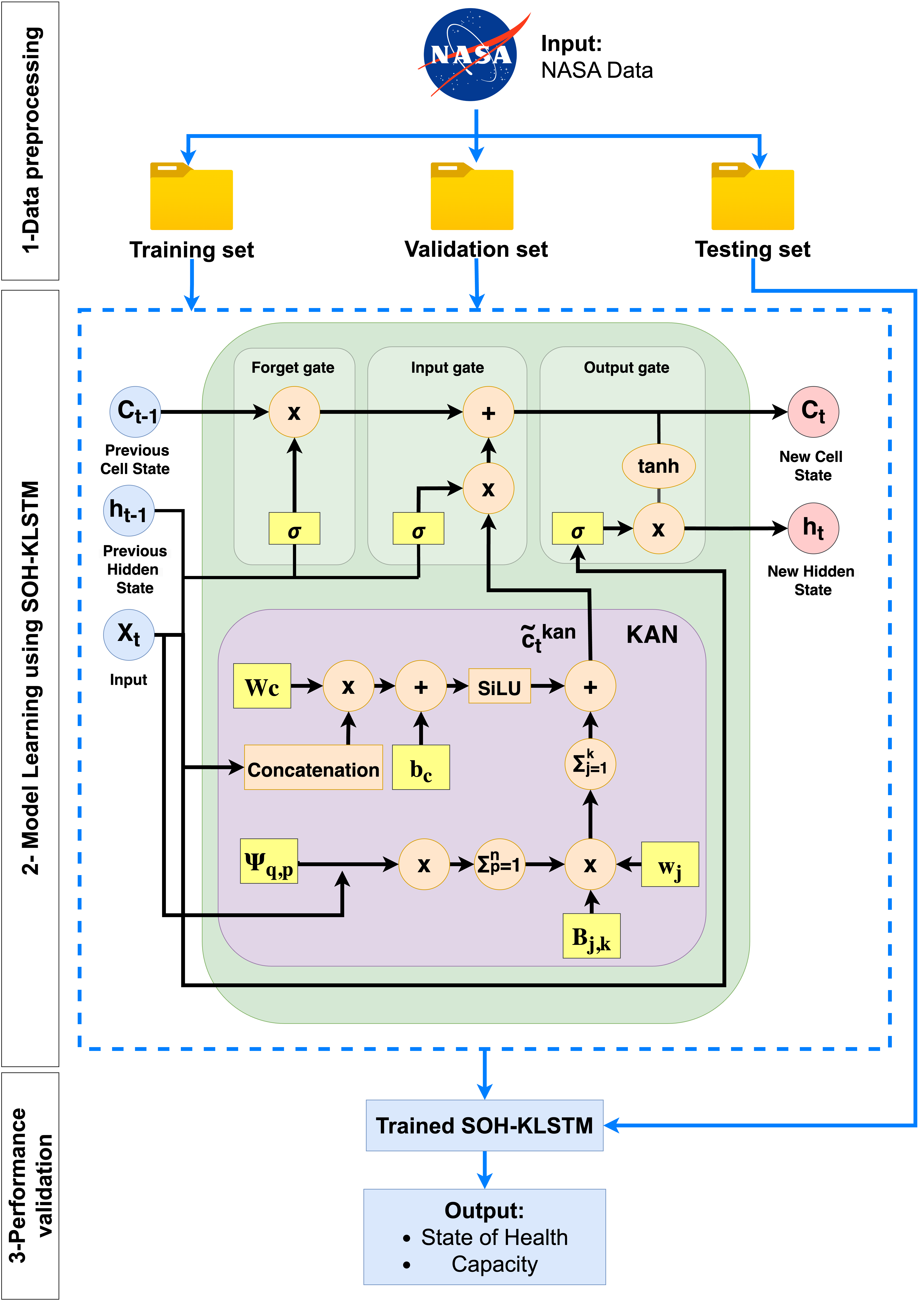}
    \caption{\textcolor{black}{The proposed SOH-KLSTM model for SOH and capacity estimation. The architecture of the SOH-KLSTM model consists of three main stages: (1) Data Preprocessing, where NASA battery datasets are split into training, validation, and testing sets; (2) Model Learning, where the input features (voltage, current, temperature, and capacity) are processed through an LSTM-based structure enhanced with a KAN for candidate cell state computation; and (3) Performance Validation, where the trained model outputs predictions for SOH and capacity.} }
    \label{fig:approach}
\end{figure*}

The proposed SOH-KLSTM model is based on the standard LSTM architecture, incorporating its essential components, including memory cells, input gates, forget gates, and output gates. The key advancement of this model lies in the refinement of the candidate cell state \(\tilde{C}_t\), which is dynamically optimized using Kolmogorov-Arnold networks. This enhancement allows the model to effectively capture both linear and highly non-linear dependencies in battery degradation trends, leading to a more accurate and reliable estimation of the SOH.  As illustrated in Figure~\ref{fig:approach}, the KAN module replaces the conventional linear transformation used in standard LSTMs with a more flexible nonlinear function mapping, generating the enhanced candidate cell state \(\tilde{C}_t^{\text{KAN}}\). Unlike traditional LSTM architectures, where candidate cell states are computed using fixed-weight transformations, KAN introduces an adaptive learning mechanism that dynamically adjusts function representations based on input variations. Specifically, KAN employs B-spline transformations (see SubSection \ref{subsection:B-Spline}) and the SiLU activation function (refer to Subsection \ref{subsection:SiLU}) to construct a robust function approximation framework. The use of B-spline transformations enables localized adaptability, allowing the model to capture fine-grained variations in SOH data, while SiLU activation ensures smooth and stable gradient propagation, improving learning efficiency.

\textcolor{black}{Once the enhanced candidate cell state \(\tilde{C}_t^{\text{KAN}}\) is calculated, it is seamlessly integrated with the input and forget gate outputs to update the final cell state \(C_t\). This adaptive integration refines the memory update process by dynamically modulating the influence of non-linear transformations, thereby capturing both short-term fluctuations and long-term degradation patterns more effectively. Using the expressive power of KAN, the model selectively preserves critical battery health information while attenuating transient noise and irrelevant perturbations. This refined update mechanism significantly bolsters the SOH-KLSTM model's capacity to extract and maintain intricate temporal dependencies from sequential SOH data. By achieving a judicious balance between memory retention and forgetting, the model improves its predictive accuracy, ensuring that essential long-term trends are learned while extraneous details are systematically discarded. Consequently, this results in improved robustness and reliability in estimating SOH and battery capacity, making the proposed approach particularly well-suited for real-world battery health monitoring applications.}

\subsection{KAN-Enhanced Candidate Cell State in LSTM}
\textcolor{black}{Traditional LSTM models compute the candidate cell state using a fixed weight transformation based on a simple linear transformation followed by tanh activation, for example. However, this conventional approach fails to capture complex non-linear dependencies in battery degradation patterns.} The KAN-Integrated Candidate Cell State in the LSTM model improves the conventional LSTM by replacing this standard linear transformation used to calculate the candidate cell state \( \tilde{C}_t \) with a more flexible non-linear transformation provided by the KAN. This enhancement aims to capture both linear trends and complex non-linear dependencies in sequential data and improve the model's predictive accuracy. \textcolor{black}{The SOH-KLSTM model is not a simple combination of two algorithms, but a fundamentally improved LSTM design that integrates KAN within the core architecture to improve predictive accuracy. This integration is achieved in an innovative and structurally unique manner.} The KAN module applies B-spline transformations and the SiLU activation function to generate the enriched candidate cell state, denoted \( \tilde{C}_t^{\text{KAN}} \). This improved candidate state incorporates more information from the input, ensuring that the model can effectively learn both simple and intricate patterns. The calculation of the KAN-enhanced candidate cell state is given by Equation \ref{eq:kan_enhanced_cell_state}:

\begin{equation}
    \tilde{C}_t^{\text{KAN}} = \text{SiLU}\left( W_C \cdot [h_{t-1}, X_t] + b_C \right) + \sum_{j=1}^{k} w_j B_{j,k}\left( \sum_{p=1}^{n} \Psi_{q,p}(X_t) \right)
    \label{eq:kan_enhanced_cell_state}
\end{equation}

where \( W_C \) is the weight matrix applied to the concatenation of the previous hidden state \( h_{t-1} \) and the current input \( X_t \). The bias term is \( b_C \), and the SiLU activation function introduces smooth nonlinearity for stable learning. The second term uses weighted B-spline transformations \( B_{j,k} \), parameterized by \( w_j \), to capture complex non-linear dependencies present in the input data.
\textcolor{black}{This novel candidate cell state refinement mechanism ensures that both short-term fluctuations and long-term degradation patterns are captured effectively, making the model more robust to non-stationary battery health dynamics.}

After computing the enriched candidate cell state \( \tilde{C}_t^{\text{KAN}} \), it is combined with the output of the input gate \( i_t \) and the forget gate \( f_t \) to generate the updated cell state using Equation \ref{eq:cell_state_update}.

\begin{equation}
    C_t = f_t \cdot C_{t-1} + i_t \cdot \tilde{C}_t^{\text{KAN}}
    \label{eq:cell_state_update}
\end{equation}

In this context, the forget gate \( f_t \) controls how much of the previous cell state \( C_{t-1} \) is retained, ensuring that relevant past information is preserved. The input gate \( i_t \) determines how much of the new candidate state \( \tilde{C}_t^{\text{KAN}} \) is added, allowing the model to effectively incorporate new information. 

\subsection{B-Spline Transformations}
\label{subsection:B-Spline}


\textcolor{black}{B-splines are piecewise polynomial functions that provide smooth and localized transformations of input features. Unlike traditional transformations, which may impose rigid functional forms, B-splines dynamically adjust to complex non-linear patterns found in battery degradation data, making them particularly suitable for SOH estimation \cite{chen2024state}. They are defined by knot points, which segment the input space and a degree \(k \) that controls the smoothness of the curve. The base case and the recursive formulation of the B-splines are expressed in the following:}

\paragraph{Base Case (Degree 0 B-spline):}

\begin{equation}
    N_{i,0}(x) = 
    \begin{cases} 
    1 & \text{if } t_i \leq x < t_{i+1}, \\
    0 & \text{otherwise}
    \end{cases}
    \label{eq:bspline_base}
\end{equation}

\paragraph{Recursive Case (Higher Degree B-splines):}
For higher-degree B-splines, the recursive formula is applied using Equation \ref{eq:bspline_recursive}:

\begin{equation}
    N_{i,k}(x) = \frac{x - t_i}{t_{i+k} - t_i} \cdot N_{i,k-1}(x) + \frac{t_{i+k+1} - x}{t_{i+k+1} - t_{i+1}} \cdot N_{i+1,k-1}(x)
    \label{eq:bspline_recursive}
\end{equation}

Where \( N_{i,k}(x) \) is the B-spline basis function of degree \( k \) at position \( i \), and \( t_i \) are the knot points that split the input space.

The transformation applied to each feature \( x_p \) in the SOH-KLSTM is represented by the \( \Psi \)-function in Equation \ref{eq:psi_function}:

\begin{equation}
    \Psi_{q,p}(x_p) = \sum_{i=1}^{k} c_{i,q,p} B_{i,k}(x_p)
    \label{eq:psi_function}
\end{equation}

where \( B_{i,k}(x_p) \) are the B-spline basis functions of degree \( k \), \( c_{i,q,p} \) are the learned coefficients, and \( m \) is the number of B-spline basis functions.

\textcolor{black}{Conventional DL models often rely on fixed activation functions that may not capture localized variations in battery degradation. In contrast, B-spline transformations provide a highly flexible, piecewise polynomial representation, and allow our SOH-KLSTM model to effectively model both gradual and abrupt changes in SOH indicators, such as voltage, current, and temperature fluctuations. By integrating learnable activations with localized B-spline transformations, KANs offer a robust and flexible approach to modeling complex high-dimensional functions, thereby overcoming the limitations of traditional MLP architectures.}

\subsection{SiLU Activation Function}
\label{subsection:SiLU}

\textcolor{black}{The Sigmoid-weighted Linear Units activation function has emerged as a compelling alternative to conventional activation functions, offering both smoothness and computational efficiency \cite{elfwing2018sigmoid}. Ensuring smooth non-linear transitions facilitates stable gradient propagation. One of the unique strengths of SiLU lies in its ability to retain negative information while preserving positive scaling, allowing neural networks to effectively capture complex patterns in high-dimensional data. This property improves the convergence and generalization of training. As a result, it has been widely adopted in various ML tasks, including image classification, object detection, and natural language processing \cite{wolff2025cvkan}, \cite{danish2025kolmogorov}. The SiLU function is formally defined as follows \cite{shuai2025physics}: }

\begin{equation}
    \text{SiLU}(x) = \frac{x}{1 + e^{-x}} = x \cdot \sigma(x),
    \label{eq:silu}
\end{equation}

where \( \sigma(x) = \frac{1}{1 + e^{-x}} \) represents the sigmoid function.\\

\textcolor{black}{The SiLU function exhibits distinct asymptotic properties for large input magnitudes:}

\textcolor{black}{\begin{itemize}
    \item For large positive inputs (\( x \to +\infty \)), the sigmoid function approaches unity:
    \begin{equation}
        \sigma(x) \to 1, \quad \text{thus, SiLU behaves as an identity function:} \quad \text{SiLU}(x) \approx x.
    \end{equation}
    \item For large negative inputs (\( x \to -\infty \)), the sigmoid function converges to zero:
    \begin{equation}
        \sigma(x) \to 0, \quad \text{thus, SiLU asymptotically approaches zero:} \quad \text{SiLU}(x) \approx 0.
    \end{equation}
\end{itemize}}
\textcolor{black}{The \(\text{SiLU}\) activation function is smooth and non-monotonic, enabling stable gradient updates and richer representations than \(\text{ReLU}\) and \(\sigma\). It is bounded below for negative inputs, yet unbounded above, effectively scaling activations and mitigating the dying ReLU problem. By allowing gradients for both positive and slightly negative inputs, SiLU enhances convergence and performance, making it ideal for our proposed SOH-KLSTM.}
\subsection{Algorithm Steps}

The SOH-KLSTM model integrates a KAN module into the LSTM architecture to predict SOH and battery capacity. The LSTM updates hidden and cell states over time, whereas the KAN module refines the candidate cell state. The input, forget and output gates regulate the information flow by retaining essential data and discarding irrelevant details. Algorithm~\ref{alg:kan_lstm} details the operations for SOH and capacity estimation. \textcolor{black}{The overall architecture of the proposed model is depicted in Figure~\ref{fig:approach}, highlighting its three key stages: data preprocessing, model learning, and performance validation. }
\begin{algorithm}[ht]
\caption{Proposed SOH-KLSTM model for SOH and Capacity Estimation}
\label{alg:kan_lstm}
\begin{algorithmic}
\STATE \textbf{Input}: Input sequence \( X_t \), hidden state \( h_{t-1} \), cell state \( C_{t-1} \)
\STATE \textbf{Output}: Predicted SOH \( \hat{y}_{\text{SOH}} \) and battery capacity \( \hat{y}_{\text{cap}} \)
\STATE \textbf{Initialization}: Initialize weights \( W_i, W_f, W_o, W_C \), KAN weights \( W_{\text{KAN}} \), recurrent weights \( U_i, U_f, U_o, U_C \), B-spline coefficients, and biases \( b_i, b_f, b_o, b_C \)
\FOR{each time step \( t \)}
    \STATE Compute pre-activation: 
    \[ z_t = W \cdot X_t + U \cdot h_{t-1} + b \]

    \STATE Split \( z_t \) into components: \( z_{t,0}, z_{t,1}, z_{t,2}, z_{t,3} \)

    \STATE Compute gate activations:
    \[ i_t = \sigma(z_{t,0}) \quad (\text{Input gate}) \]
    \[ f_t = \sigma(z_{t,1}) \quad (\text{Forget gate}) \]
    \[ o_t = \sigma(z_{t,3}) \quad (\text{Output gate}) \]

    \STATE Compute KAN-enhanced candidate cell state:
    \[ \tilde{C}_t^{\text{KAN}} = \text{SiLU}(W_{\text{KAN}} \cdot z_{t,2} + b_{\text{KAN}}) + \sum_{i=1}^{k} w_i B_{i,k}(X_t) \left( \sum_{p=1}^{n} \Psi_{q,p}(X_t) \right) \]

    \STATE Update the cell state:
    \[ C_t = f_t \cdot C_{t-1} + i_t \cdot \tilde{C}_t^{\text{KAN}} \]

    \STATE Update the hidden state:
    \[ h_t = o_t \cdot \tanh(C_t) \]
    
\ENDFOR
\STATE \textbf{Final SOH and Capacity Estimation}:
\[
\hat{y}_{\text{SOH}}, \hat{y}_{\text{cap}} = W_{\text{out}} \cdot h_t + b_{\text{out}}
\]
\STATE \textbf{Return}: Predicted SOH \( \hat{y}_{\text{SOH}} \) and battery capacity \( \hat{y}_{\text{cap}} \)
\end{algorithmic}
\end{algorithm}
\section{Experiments and Results}
\label{sec:results}
This section provides a detailed overview of the dataset, experimental setup, and results analysis of the SOH-KLSTM model to predict the SOH of Li-ion batteries.

\subsection{Dataset}

For the evaluation of the SOH-KLSTN model, we have used several subsets from NASA’s Prognostics Center of Excellence (PCoE) Battery Dataset, which contains data from 34 lithium-ion (Li-ion) 18650 cells, each with a capacity of 2 Ah. These cells were cycled to 70\% or 80\% of their original capacity under various temperature conditions using a custom-built battery tester. The cycling process included three key phases: charging, discharging, and electrochemical impedance spectroscopy (EIS).

The charging was carried out using a constant current-constant voltage (CC-CV) method at 1.5 A until the cells reached 4.2 V, with a cutoff current of 20 mA. Various discharge profiles were used to simulate realistic use and accelerate degradation. EIS was conducted with a frequency sweep from 0.1 to 5 KHz, providing detailed information on the internal electrochemical properties of the cells. This dataset captures valuable battery performance and degradation patterns under various operational conditions.

The selected subsets include data from rechargeable Li-ion 18650 batteries, specifically: B0005 (B05), B0007 (B07), B18, B33, B34, B46, B47, and B48. Each subset includes different operational and environmental conditions, offering a comprehensive basis for evaluating battery health through various stress factors. This diversity is essential for building a robust model capable of accurately predicting SOH in different usage scenarios. \textcolor{black}{For enhanced analysis, the datasets have been organized based on uniform condition datasets, current discharge conditions, and temperature profiles, as follows \cite{zheng2019state}, \cite{lim2023optimal}, \cite{chen2024contribution}:}

\begin{enumerate}
    \item \textbf{Group A: Uniform Condition Datasets at Ambient Temperature} \\
    This group includes batteries B05, B07, and B18, which were tested under identical conditions with a discharge current of 2A and an ambient temperature of 24°C. These datasets simulate moderate operating conditions, providing a baseline for comparison with more demanding scenarios. They serve as a control group to analyze battery degradation under normal environmental conditions.

    \item \textbf{Group B: High-Power Cycle Datasets} \\
    This group includes batteries B33 and B34, tested with a high discharge current of 4A. These datasets represent high-power applications where frequent rapid charging and discharging cycles accelerate aging due to thermal stress. They are essential for evaluating the durability of Li-ion batteries in demanding applications, such as electric vehicles. 

    \item \textbf{Group C: Low-Temperature Datasets} \\ 
    Group C consists of batteries B46, B47, and B48, tested at a low temperature of 4°C with various discharge currents. These datasets help assess the effects of cold environments on battery performance, including increased internal resistance, reduced charge retention, and overall efficiency loss. This information is crucial for applications requiring battery reliability in extreme cold, such as outdoor power systems and cold-climate operations.
\end{enumerate}

\subsection{Experimental Setup}
The experiments were carried out on a machine equipped with an Intel Core i9-11900K processor (3.50 GHz, 8 cores, 16 threads), 64GB of DDR4 RAM, and an NVIDIA RTX 3080 GPU with 10GB of VRAM, running Ubuntu 20.04 LTS. The implementation was performed using Python 3.8, with key libraries including TensorFlow 2.x and Keras for constructing and training the neural network. Data preprocessing and manipulation were managed using NumPy and Pandas, while SciPy facilitated mathematical operations such as B-spline transformations to enhance feature extraction.

For performance evaluation, scikit-learn was employed to compute metrics such as RMSE and execution time. Data visualization and further analysis were performed using Matplotlib, ensuring that both qualitative and quantitative aspects of the model performance were thoroughly assessed.

Detailed steps for the implementation of the SOH-KLSTM model are outlined below. 
\begin{itemize}
    \item \textbf{Data Preprocessing:}  
    The dataset is divided into subsets of training (70\%), validation (20\%) and tests (10\%) to guarantee a reliable generalization. A MinMaxScaler is used to normalize the data and scale all features to the same range. This step improves model convergence and prevents features with larger numerical values from dominating the learning process.

    \item \textbf{Model Learning:}  
    The training process is performed using the Adam optimizer with a learning rate of 0.001, a batch size of 32, and a maximum of 100 epochs per fold. Early stopping is implemented to halt training if the validation loss does not improve for 10 consecutive epochs. Adaptive learning rate reduction further refines the learning process by adjusting the learning rate when validation loss plateaus, ensuring optimal convergence.

    \item \textbf{Performance Validation:}  
    The model’s performance is evaluated using RMSE for SOH predictions, providing a measure of prediction accuracy. The execution time for each fold is also recorded to assess computational efficiency, confirming the practicality of the model for real-world applications.
\end{itemize}

\subsection{Eperimental Results}

\begin{figure*}[h!]
\centering
\includegraphics[width=0.9\textwidth]{ 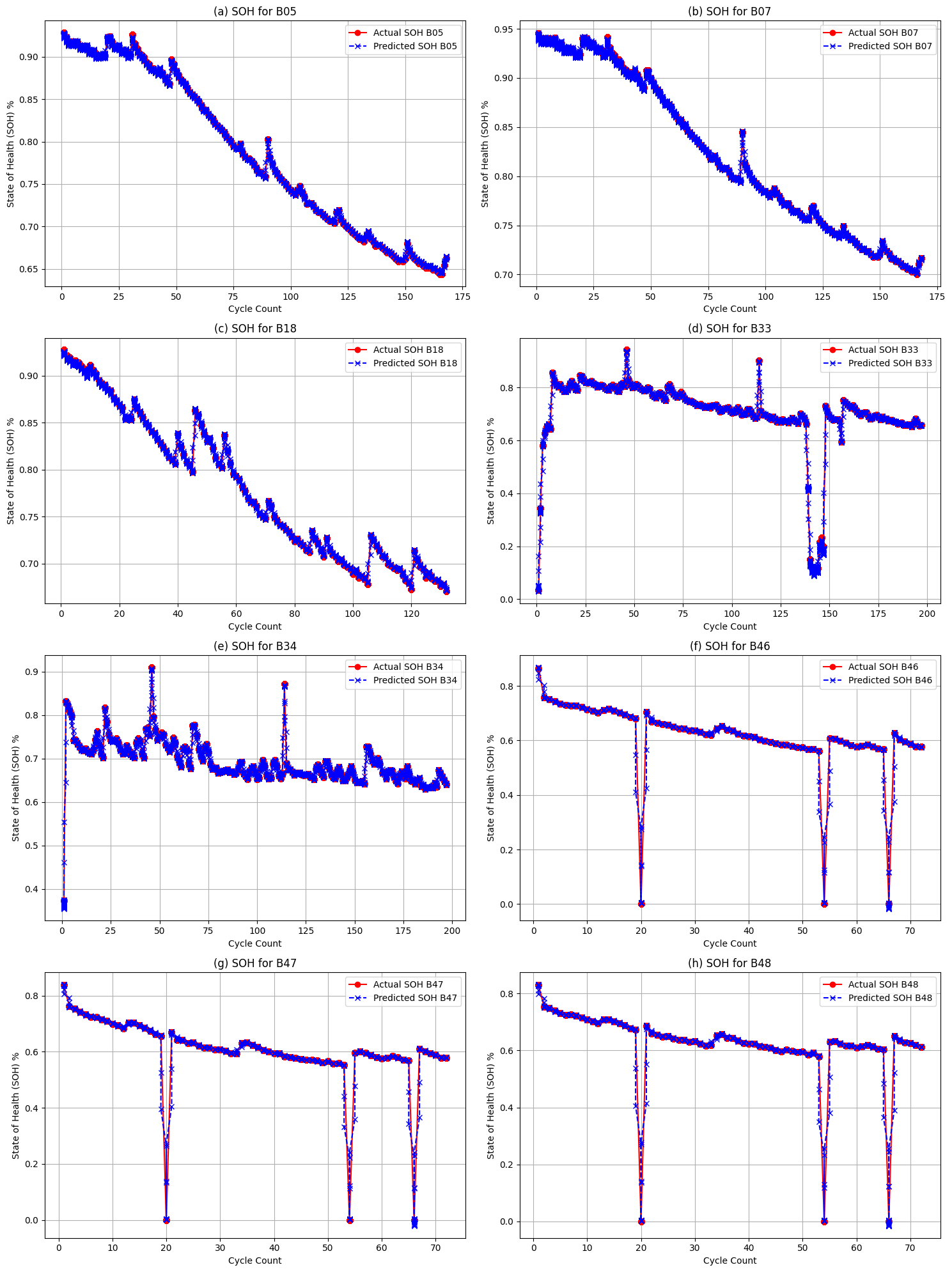}
\caption{
\textcolor{black}{Comparison of actual and predicted SOH for different lithium-ion battery datasets: (a) B05, (b) B07, (c) B18, (d) B33, (e) B34, (f) B46, (g) B47, and (h) B48. The plots show the actual SOH (red points) and the predicted SOH (blue points) over battery cycle counts. } }
\label{fig:soh_comparison}
\end{figure*}

The SOH-KLSTM model achieved high predictive accuracy in all subsets of battery data tested. Figure~\ref{fig:soh_comparison} compares the actual and predicted SOH for various battery cells. The SOH-KLSTM model effectively tracks the actual SOH values, even when batteries exhibit irregular degradation patterns. This performance demonstrates the model’s ability to capture both continuous and nonlinear degradation behaviors, offering reliable SOH predictions over multiple cycles.
Table~\ref{tab1} presents the evaluation metrics, with RMSE quantifying predictive accuracy and execution time reflecting computational efficiency.

\subsubsection{RMSE Analysis}



    

\textcolor{black}{The RMSE values presented in Table~\ref{tab1} highlight the high accuracy of the SOH-KLSTM model across all datasets and demonstrate its robustness under different operating conditions, including varying temperatures and discharge rates. To further analyze the adaptability of the model, we evaluate its performance under moderate, high-power, and low-temperature conditions, providing insights into its applicability in the real world in various scenarios.}

\textcolor{black}{
For moderate conditions (Group A: B05, B07, B18), tested at an ambient temperature of 24°C with a discharge current of 2A, the model achieves impressively low RMSE values: 0.001682 for B05, 0.002112 for B07 and 0.001816 for B18. These results confirm SOH-KLSTM's high predictive accuracy in stable environments, making it well-suited for applications such as consumer electronics and renewable energy storage systems, where temperature fluctuations are minimal and battery degradation follows predictable trends.}

\textcolor{black}{
For high-power cycle conditions (Group B: B33, B34), batteries were subjected to higher discharge currents (4A), making them representative of electric vehicle and industrial applications where batteries experience frequent rapid charging and discharging. The model maintains strong performance with RMSE values of 0.003976 for B33 and 0.002342 for B34. The slight increase in RMSE compared to moderate conditions is attributed to the stress imposed by higher discharge rates and temperature variations. B33 and B34 were tested at a low temperature of 4°C, in contrast to the 24°C condition of Group A. At 4°C, battery efficiency typically decreases due to increased internal resistance and reduced ion mobility. The SOH-KLSTM model successfully captures battery degradation trends in these challenging conditions, confirming its stability and effectiveness for high-performance applications such as electric vehicles, where rapid charge-discharge cycles and temperature shifts significantly affect battery health.}

\textcolor{black}{
Batteries B46, B47, and B48 were tested under similar low-temperature conditions (approximately 4°C) but at a much lower discharge rate of 1A. Here, the RMSE values were slightly higher, with 0.006440 for B46, 0.006888 for B47, and 0.007114 for B48. Although these errors are marginally higher than those in Group B, they remain within an acceptable range. This suggests that while lower discharge currents in extremely cold conditions may introduce a bit more variability (likely due to the combined effects of reduced ion mobility and the less aggressive cycling), the model still robustly tracks the battery's state of health.}

\textcolor{black}{The results confirm that SOH-KLSTM is highly adaptable to a wide range of operating conditions. Although the model achieves the lowest RMSE in moderate conditions, it also shows strong resilience in high-power cycles and low-temperature environments, making it suitable for real-world SOH monitoring in diverse applications.}

\begin{table*}[h!]
\centering
\caption{Execution Time and RMSE for SOH-KLSTM on Various Data Subsets}
\vspace{0.1cm}
\label{tab1}
\begin{tabular}{@{}lcc@{}}
\toprule
Dataset & RMSE & Execution Time (s) \\
\midrule
B05  & 0.001682 & 2.627 \\
B07  & 0.002112 & 2.506 \\
B18  & 0.001816 & 1.736 \\
B33  & 0.003976 & 3.422 \\
B34  & 0.002342 & 2.380 \\
B46  & 0.006440 & 1.260 \\
B47  & 0.006888 & 1.316 \\
B48  & 0.007114 & 2.614 \\
\bottomrule
\end{tabular}
\footnotetext{Source: Experimental results for SOH-KLSTM model on NASA’s Prognostics Center of Excellence datasets.}
\end{table*}

\subsubsection{Execution Time Analysis}

The computational efficiency of the SOH-KLSTM model was assessed based on execution time, as presented in Table~\ref{tab1}. The model consistently achieved fast processing times, demonstrating its suitability for real-time battery management, where timely and accurate SOH predictions are essential. Execution times ranged from 1.260 s (B46) to 3.422 s (B33), highlighting the model’s efficiency even in high-power datasets. This rapid processing capability ensures continuous monitoring and real-time decision-making, making SOH-KLSTM highly applicable in battery health diagnostics and predictive maintenance.

\subsection{Comparison with State-of-The-Art Methods}

\textcolor{black}{To evaluate the effectiveness of the proposed SOH-KLSTM model, we compare its performance against several state-of-the-art SOH estimation models, including LSTM \cite{yao2023novel}, CNN-LSTM \cite{zhu2023state}, SVR-LSTM \cite{obisakin2022state}, DEGWO-LSTM \cite{ma2022novel}, GPR \cite{yao2024state}, SBL \cite{li2023lithium}, CMMOG \cite{zhang2025cmmog}, and DGL-STFA \cite{chen2025dgl}, as summarized in Table \ref{table:rmse_comparison}. The comparison is based on RMSE and Mean Absolute Percentage Error (MAPE), along with percentage error reduction, providing a comprehensive assessment of predictive accuracy and model efficiency. RMSE quantifies large error magnitudes, while MAPE offers a relative measure of prediction accuracy, which is particularly relevant for real-world SOH estimation applications.}


\textcolor{black}{On the B0005 dataset, SOH-KLSTM achieves the lowest RMSE of 0.001682, representing a 97.12\% error reduction compared to LSTM (0.058334), while on the B0007 dataset, it achieves 0.002112 RMSE, marking a 94.85\% reduction over LSTM (0.041061). Furthermore, SOH-KLSTM demonstrates superior accuracy, with a MAPE of 0.17\% on B0005, significantly lower than CNN-LSTM (2.00\%) and LSTM (5.83\%). Compared to CNN-LSTM, SOH-KLSTM achieves a 65.74\% improvement in RMSE due to its enhanced ability to capture long-term dependencies. Compared to SVR-LSTM, which records an RMSE of 0.003, SOH-KLSTM performs better, benefitting from the KAN framework’s superior nonlinear approximation capability. These results confirm the superior accuracy and efficiency of SOH-KLSTM to predict battery SOH, which makes it highly suitable for real-world applications.}

\textcolor{black}{To further assess the effectiveness of SOH-KLSTM, we compared it with the Gaussian Process Regression (GPR) model proposed by Yao et al. \cite{yao2024state}. This model demonstrates strong theoretical performance, but is highly sensitive to the variability of training data, which reduces its reliability in dynamic environments.}

\textcolor{black}{Additionally, we compared SOH-KLSTM with the LSTM-based SOH estimation model that incorporates health indicator selection and hyperparameter optimization using the Differential Evolution Grey Wolf Optimizer (DEGWO) \cite{ma2022novel}. This method selects voltage, current, and temperature-based HIs, applies Pearson correlation and Neighborhood Component Analysis (NCA) to eliminate redundancy, and optimizes LSTM hyperparameters using DEGWO. Although DEGWO-LSTM achieves improved feature selection, it falls short in terms of nonlinear feature representation and temporal modeling, resulting in higher RMSE values (0.325 on B0005 and 0.377 on B0007) compared to SOH-KLSTM. }

\textcolor{black}{Furthermore, we evaluated SOH-KLSTM against the Sparse Bayesian Learning (SBL) model proposed by Li et al. \cite{li2023lithium}, which extracts multi-source HIs from voltage, temperature, and incremental capacity curves. The SBL framework employs Bayesian inference and feature selection using Pearson correlation to improve estimation accuracy. However, it imposes high computational costs, which reduces its real-time applicability. The SBL model records high RMSE values (3.5656 on B0005 and 2.6153 on B0007), indicating a weaker predictive precision compared to SOH-KLSTM.}

\textcolor{black}{We also compared our model with the Convolutional Neural Network-Multi-gate Mixture of Gated Recurrent Units (CMMOG) model proposed by Zhang et al. \cite{zhang2025cmmog}. This approach integrates multi-task learning for simultaneous SOH regression across multiple battery conditions, combining CNN for feature extraction, GRU for state mapping, and a multi-gated network for weight optimization. However, it fails to maintain high precision across different charge/discharge conditions, with RMSE values as high as 0.6490 on B0005.  }

\textcolor{black}{Finally, we compare SOH-KLSTM with the DGL-STFA model \cite{chen2025dgl}, which leverages dynamic graph learning and spatial–temporal fusion attention mechanisms to model evolving relationships between battery health indicators. DGL-STFA records an RMSE of 0.876, highlighting its computational inefficiency compared to the proposed SOH-KLSTM.}

\textcolor{black}{These results confirm that SOH-KLSTM achieves state-of-the-art accuracy while maintaining computational efficiency, making it an optimal solution for real-world applications, including electric vehicles and industrial battery management systems.}

\begin{table*}[h]
\caption{Comparison of proposed SOH-KLSTM performance for SOH prediction with other models proposed in the literature.}
\label{table:rmse_comparison}
\begin{tabular*}{\textwidth}{@{\extracolsep\fill}lccccccc}
\toprule%
& \multicolumn{3}{@{}c@{}}{B0005 Dataset\footnotemark[1]} & \multicolumn{4}{@{}c@{}}{B0007 Dataset\footnotemark[2]} \\\cmidrule{2-5}\cmidrule{6-8}%
Ref & Model & \makecell{RMSE \\ $\downarrow$(\%)} & \makecell{MAPE \\ $\downarrow$(\%)}& \makecell{Error \\ Reduction $\uparrow$(\%
)} & \makecell{RMSE  \\$\downarrow$(\%)} & \makecell{MAPE\\ $\downarrow$(\%)} & \makecell{Error \\Reduction $\uparrow$(\%) }\\
\midrule
\makecell[l]{Obisakin et al. \\\cite{obisakin2022state}}& SVR-LSTM   & 0.003  & 0.30  & 94.86\%     & --       & --      & --\\
\textcolor{black}{Ma et al.} \cite{ma2022novel}  & \textcolor{black}{DEGWO-LSTM} & \textcolor{black}{0.325}  &\textcolor{black}{0.467} & -  & \textcolor{black}{0.377}  & \textcolor{black}{0.423}  & - \\
Yao et al. \cite{yao2023novel}& LSTM  & 0.058334 & 5.83  & --           & 0.041061 & 4.11  & --\\
Zhu et al.\cite{zhu2023state} & CNN-LSTM  & 0.02  & 2.00  & 65.74     & 0.03     & 3.00  & 26.83 \\
\textcolor{black}{Lie et al.} \cite{li2023lithium}  & \textcolor{black}{SBL } & \textcolor{black}{3.5656 }  &\textcolor{black}{- } & -  & \textcolor{black}{ 2.6153 }  & \textcolor{black}{- }  & - \\
\textcolor{black}{Yao et al.} \cite{yao2024state}  & \textcolor{black}{GPR} & \textcolor{black}{0.0114}  &  - & -  & -  & -  & - \\
\textcolor{black}{Zhang et al. \cite{zhang2025cmmog}}  & \textcolor{black}{CMMOG} & \textcolor{black}{0.6490}  &\textcolor{black}{0.4731} & \textcolor{black}{-}  & \textcolor{black}{  }  & \textcolor{black}{- }  & \textcolor{black}{-} \\

\textcolor{black}{Chen et al. \cite{chen2025dgl}}  & \textcolor{black}{DGL-STFA } & \textcolor{black}{0.876}  &\textcolor{black}{-} & \textcolor{black}{-}  & \textcolor{black}{0.876}  & \textcolor{black}{-}  & \textcolor{black}{-} \\
\textbf{Proposed model} &\textbf{SOH-KLSTM} & \textbf{0.001682} & \textbf{0.17}  & \textbf{97.12}     & \textbf{0.002112} & \textbf{0.21}  & \textbf{94.85} \\

\hline
\end{tabular*}
\end{table*}

\subsection{Discussion}
The results confirm that the SOH-KLSTM model provides an optimal balance between accuracy and efficiency. The low RMSE values across diverse datasets highlight the model’s ability to accurately predict SOH under various conditions, including challenging high-power and low-temperature environments. In addition, its fast execution times support real-time applications, such as electric vehicles, consumer electronics, and grid energy storage. The SOH-KLSTM model sets a new standard for the estimation of SOH in Li batteries, providing a reliable, scalable, and real-time solution for accurate battery health prediction.

Below, we highlight the major findings, related to the proposed SOH-KLSTM, from the experiments:

\begin{itemize}
    \item \textbf{Predictive Accuracy:} The SOH-KLSTM model consistently achieves lower RMSE values across datasets, outperforming conventional and hybrid methods. The inclusion of KAN improves the model’s ability to capture intricate degradation patterns, leading to superior prediction accuracy.
    \item \textbf{Generalization Across Operating Conditions:} The SOH-KLSTM model demonstrates robustness across a range of operational conditions, including high-power cycles and low-temperature environments. For example, on the B33  high-power cycle dataset, the model achieved an RMSE of 0.003976, while on the low-temperature B46 dataset, the RMSE was 0.006440, confirming its versatility for real-world applications such as electric vehicles.
    \item \textbf{Computational Efficiency:} With execution times ranging between \textbf{1.26 s} and \textbf{3.42 s}, the SOH-KLSTM model is computationally efficient, making it well-suited for real-time applications. Fast execution is particularly important for BMS, where timely SOH predictions are essential.
\end{itemize}

The integration of KAN into the LSTM architecture significantly enhances the ability of the model to capture the complex, non-linear dynamics present in battery degradation processes. These key improvements distinguish the SOH-KLSTM model from conventional LSTM approaches:

\begin{itemize}
    \item \textbf{Non-Linear Representation}: SOH-KLSTM utilizes B-spline transformations for each input feature, enabling the model to capture localized non-linear dependencies that traditional LSTM models often fail to grasp.
    
    \item \textbf{Localized Feature Interactions}: By incorporating B-spline basis functions, SOH-KLSTM achieves fine-grained control over input-output mappings. This is crucial for the estimation of SOH, as battery degradation patterns vary between different regions of the input space.
    
    \item \textbf{SiLU Activation for Stability}: The inclusion of the SiLU activation function ensures smooth, stable learning by maintaining zero-mean and unit-variance activations, accelerating convergence and mitigating issues like vanishing or exploding gradients.
    
    \item \textbf{Improved Prediction Accuracy}: By combining B-spline transformations for local modeling with SiLU’s stabilizing effects, SOH-KLSTM enhances prediction accuracy for SOH and battery capacity, capturing both global trends and localized patterns for more reliable predictions.
\end{itemize}

\section{Conclusion}
\label{sec:conclusion}
The SOH-KLSTM model offers significant advances in the prediction of SOH for lithium-ion batteries by integrating KAN to enhance the candidate cell state within LSTM networks. This integration, combined with experimental validation on multiple subsets of NASA’s Prognostics Center of Excellence (PCoE) dataset, demonstrates that SOH-KLSTM achieves superior predictive accuracy, with up to a 97.12\% reduction in prediction error compared to baseline models. The powerful ability of the KAN module to capture complex and non-linear degradation patterns further boosts the model's performance, enabling it to outperform advanced methods such as CNN-LSTM and SVR-LSTM. Consequently, the SOH-KLSTM model provides reliable SOH predictions in diverse operational conditions, including high-power cycles and low-temperature environments, making it ideal for applications in electric vehicles, renewable energy storage systems, and other industries requiring efficient battery management. \textcolor{black}{Future work will focus on predicting Remaining Useful Life (RUL), improving computational efficiency, and extending SOH-KLSTM to battery packs. By applying the model at the cell level and aggregating predictions using fusion strategies (e.g., weighted averaging, and adaptive techniques), our aim is to develop a robust framework for pack-level SOH estimation, enhancing its real-world applicability.}

\section*{Declaration of interests}

The authors declare that they have no known financial interests or personal relationships that could have influenced the work presented in this paper.
\section*{Acknowledgements}
The authors would like to thank Prince Sultan University for their support.
\printcredits

\bibliographystyle{unsrt}

\bibliography{main_revised}



\end{document}